\documentclass[sigconf]{acmart}
\AtBeginDocument{%
  \providecommand\BibTeX{{%
    \normalfont B\kern-0.5em{\scshape i\kern-0.25em b}\kern-0.8em\TeX}}}

\setcopyright{acmcopyright}
\copyrightyear{2023}
\acmYear{2023}
\acmDOI{XXXXXXX.XXXXXXX}

\acmConference[ACMMM '23]{ACM Multimedia}{October 29--November 03, 2023}{Ottawa, Canada}
\acmPrice{15.00}
\acmISBN{978-1-4503-XXXX-X/18/06}

\begin{document}

\title{4DSR-GCN: 4D Video Point Cloud Upsampling using\\ Graph Convolutional Networks}

\author{Lorenzo Berlincioni}
\email{lorenzo.berlincioni@unifi.it}
\affiliation{%
  \institution{MICC, Università degli Studi di Firenze}
  \country{Italy}
}
\author{Stefano Berretti}
\email{stefano.berretti@unifi.it}
\affiliation{%
  \institution{MICC, Università degli Studi di Firenze}
  \country{Italy}
  }
  
\author{Marco Bertini}
\email{marco.bertini@unifi.it}
\affiliation{%
  \institution{MICC, Università degli Studi di Firenze}
  \country{Italy}
  }

\author{Alberto Del Bimbo}
\email{alberto.delbimbo@unifi.it}
\affiliation{%
  \institution{MICC, Università degli Studi di Firenze}
  \country{Italy} }

\renewcommand{\shortauthors}{Anonymous Authors}
\newcommand{\etal}{\textit{et al.}}
\newcommand{\etc}{\textit{etc.}}
\newcommand{\ie}{\textit{i.e.}}
\newcommand{\eg}{\textit{e.g.}}

\begin{abstract}
Time varying sequences of 3D point clouds, or \textit{4D} point clouds, are now being acquired at an increasing pace in several applications (\eg, LiDAR in autonomous or assisted driving). In many cases, such volume of data is transmitted, thus requiring that proper compression tools are applied to either reduce the resolution or the bandwidth. 
In this paper, we propose a new solution for upscaling and restoration of time-varying 3D video point clouds after they have been heavily compressed. 
Our model consists of a specifically designed Graph Convolutional Network (GCN) that combines Dynamic Edge Convolution 
and Graph Attention Networks for feature aggregation 
in a Generative Adversarial setting. 
We present a different way to sample dense point clouds with the intent to make these modules work in synergy to provide each node enough features about its neighbourhood in order to later on generate new vertices.
Compared to other solutions in the literature that address the same task, our proposed model is capable of obtaining comparable results in terms of quality of the reconstruction, while using a substantially lower number of parameters ($\simeq$ 300KB), making our solution deployable in edge computing devices such as LiDAR.
\end{abstract}



\begin{teaserfigure}
  \includegraphics[width=\textwidth]{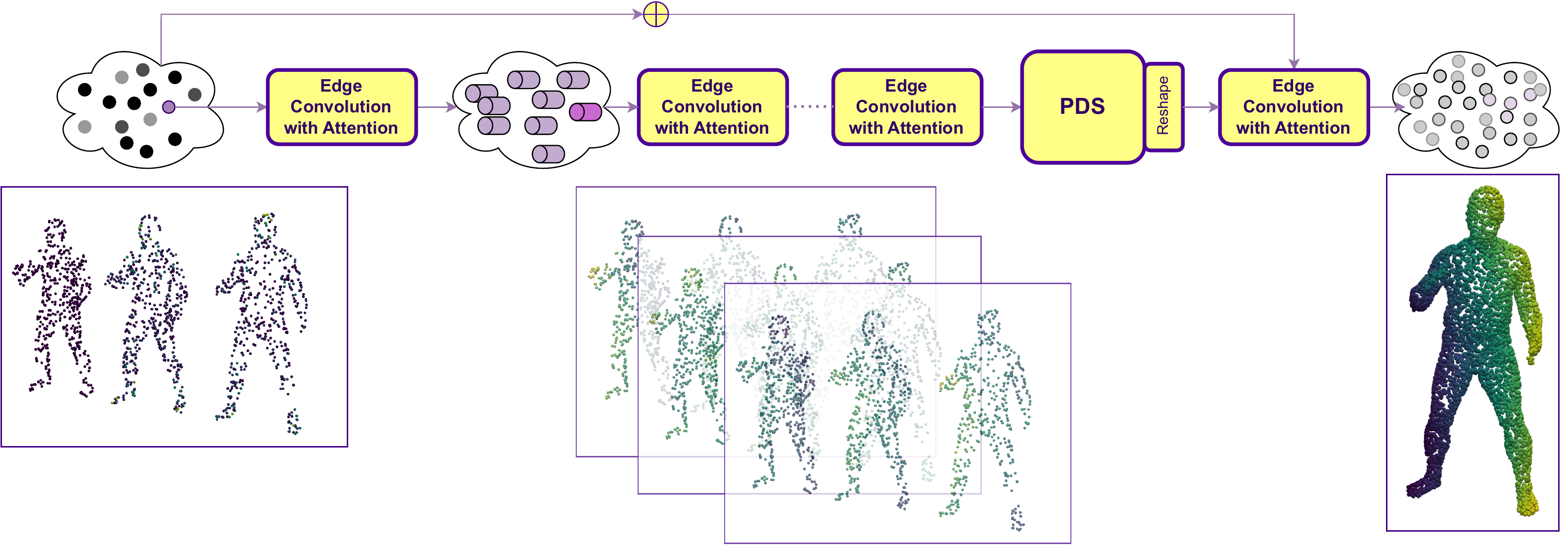}
  \caption{\textit{Top}: Schematic architecture of the Generator GCN used for our upsampling model. A set of Edge Convolution with Attention layers are cascaded to generate an embedding from the input point clouds; the Parallel Double Sampling (PDS) layer is then used for upsampling, and the output is summed at the end with the input following a residual-like schema. \textit{Bottom}: A short set of point cloud frames used as input are shown on the left; then, the output of the intermediate Edge Convolution with Attention layers are shown in the middle; the upsampled output point cloud is given on the right.}
  \label{fig:teaser}
\end{teaserfigure}

\received{20 February 2007}
\received[revised]{12 March 2009}
\received[accepted]{5 June 2009}

\maketitle

\section{Introduction}
In light of emerging applications such as Augmented and Virtual Reality (AR/VR), there is a rising interest in capturing the real world in 3D at high-resolution. For real time applications in dynamic settings, such as 3D sensing for robotics, telepresence, automated driving applications using LiDAR, this technology might need high-resolution point clouds with up to millions of points per frame.
After taking into consideration the average point-cloud video, under some constraints such as keeping the identity of a human subject recognizable, we observe that the size of a single instance, which is a single frame, can be approximated as $\sim$10 Mbytes, which translates to a bitrate of $\sim$300 Mbytes per second without compression for a 30-fps dynamic point cloud. The high data rate is one of the main problems faced by dynamic point clouds, and efficient compression technologies to allow for the distribution of such content are still widely sought. One result in this direction, is represented by the Point Cloud Compression standard specifications that include video-based PCC (V-PCC) and geometry-based PCC (G-PCC)~\cite{graziosi_nakagami_kuma_zaghetto_suzuki_tabatabai_2020} as released in 2020 by the The Moving Picture Expert Group (MPEG).

\begin{figure}[ht!]
\centering
\includegraphics[width=\linewidth]{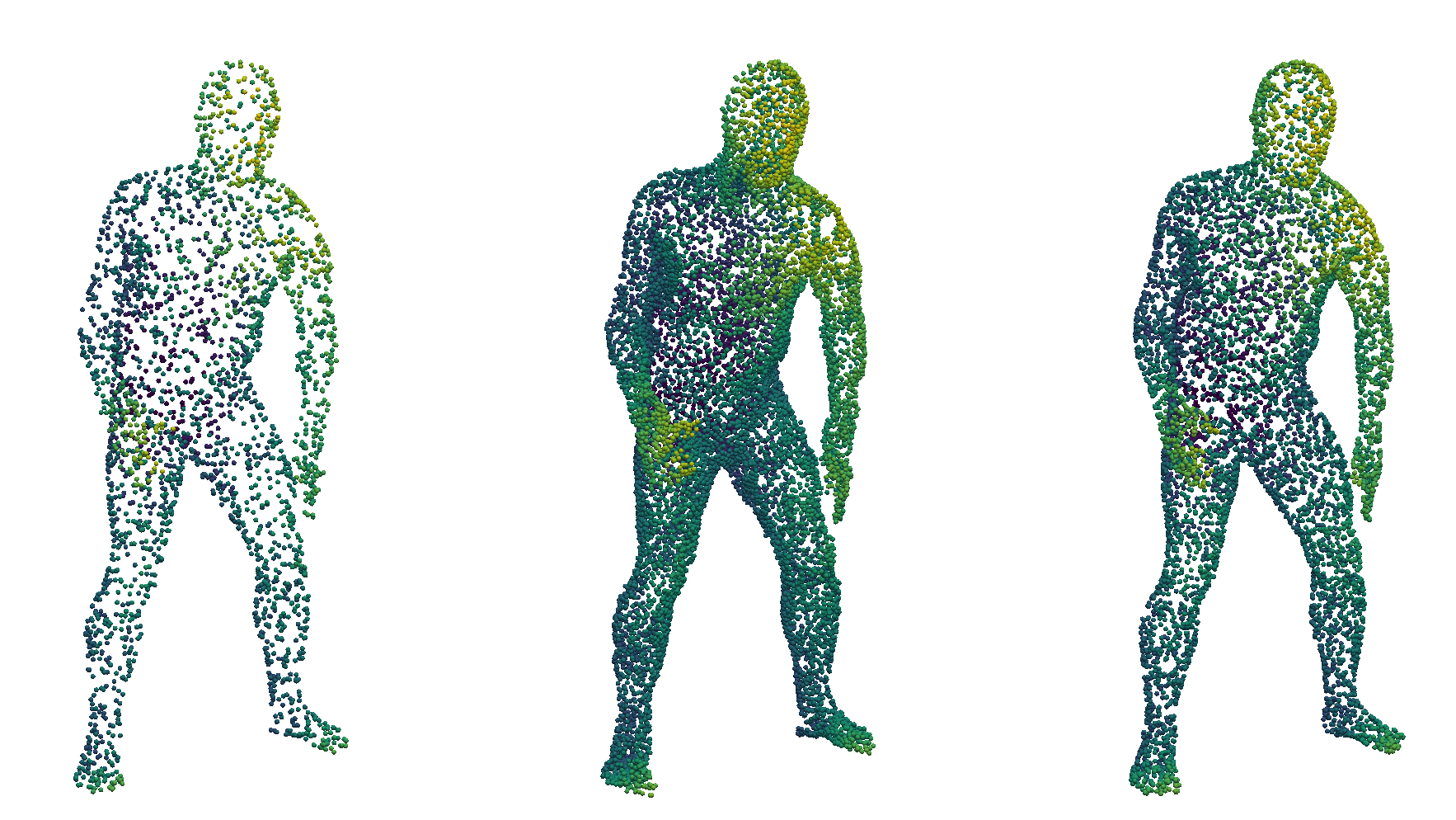}
\caption{\textit{Left:} Sample of an input low-resolution point cloud with $\sim$3K vertices. \textit{Center:} our model reconstruction with $\sim$12K vertices. \textit{Right:} Ground truth point cloud with $\sim$12K vertices.}
\label{fig:pointcloudcompression}
\end{figure}

Given these premises, our task is to perform \textit{upscaling} and \textit{artifact removal} of sparsely populated 
3D point cloud videos. 
The terms \textit{upscaling} and \textit{artifact removal} are usually found in image/video super resolution literature and so they might not have an immediate translation in the 3D context.
We will use the term \textit{upscale} to indicate the operation by which the total number of vertices of an input point cloud is increased; by using \textit{artifact removal}, instead, we will imply the correct reconstruction process after some sort of compression or subsampling has been performed on an input point cloud. 
As shown in Figure~\ref{fig:pointcloudcompression}, a high compression rate can achieve acceptable bandwidth requirements with a huge decrease in fidelity. For some applications, for example where the user experience is important, the identity of the subject must be maintained or, for autonomous driving, such a low-resolution is not acceptable.

Recent approaches that tackled this task, such as~\cite{tavu2022rfnet4d}, employed a strategy that uses long sequences of input frames and a large encoder-decoder model. As we will detail below, we followed a different approach. 

In this paper, we pose the upscaling problem in a Generative Adversarial setting using two architectural modules from the literature: the EdgeConvolution~\cite{wang2019dynamic} and the Graph Attention Network (GAT)~\cite{velivckovic2017graph}. 
In particular, the input point clouds are modeled as graphs and processed by a Graph Convolutional Network (GCN). The convolution operation has been performed using a EdgeConv module: this module incorporates local neighborhood information, can be stacked to learn global shape properties, and affinity in feature space captures semantic characteristics over potentially long distances in the original embedding. 
While this module was used for CNN-based high-level tasks on point clouds, including classification and segmentation, the GAT has been used for feature aggregation performing an attentioned learned mean of the neighbourhood features instead of simply averaging it out.
Experiments have been performed on the FAUST 4D dataset~\cite{dfaust:CVPR:2017} also in comparison with state-of-the-art solutions. Overall, our method shown upscaling reconstructions that are comparable with those reported in the literature, while using a lower number of input frames and an architecture with a much lower number of parameters. This opens the way to the deployment of our architecture in edge computing devices.

In summary, the main contributions of our work can be summarized as follows:
\begin{itemize}
\item We propose a new architecture for time varying point cloud upscaling that combines together a PointNet~\cite{pointnet++}, used as a Discriminator Network in a GAN, and a Generator that makes use of Edge Convolution on the input graphs derived from the point clouds and a graph attention mechanism for aggregating the features of the local neighbourhood. The resulting GAN architecture represents a setting that, to our knowledge, has not been tried before for this task;
\item The proposed solution demonstrates a clear advantage over the existing methods in the capability of producing upscaled 3D point clouds with comparable accuracy but using a way lower number of parameters in the architecture. 
Finally, the inference time is compatible with an online application of the method handling a stream of input frames.
\end{itemize}

\section{Related work}
Numerous studies have been conducted with the goal of reconstructing a 3D model given inputs in various possible forms: a mesh, a 3D point cloud, a collection of voxels or an implicit function. Some of these works focused on the use of a 3D point cloud as an input~\cite{Fan_2017_CVPR, frustum2017}. Others, instead, used a discretized version based on \textit{voxels}, such as~\cite{Girdhar16b, voxels45}, or directly tried to reconstruct a mesh~\cite{deepcubesmarching, wang2018pixel2mesh}.

Point cloud upsampling was first approached using optimization based solutions, while deep learning based methods were applied only more recently. Methods from both these categories are summarized below.

\smallskip

\textbf{Optimization-based methods}. 
One of the first work addressing point sets upsampling was proposed by Alexa~\etal~\cite{Alexa-TVCG:2003}. In their approach, points at vertices of a Voronoi diagram were interpolated in the local tangent space. 
Lipman~\etal~\cite{Lipman-TOG:2007}, presented a Locally Optimal Projection (LOP) operator performing points resampling and surface reconstruction using L1-median. The LOP operator showed satisfactory results even in the case the input point set was affected by noise and outliers. 
An improved version of the LOP approach aiming to address the density problem of the upscaled point set was then proposed by Huang~\etal~\cite{Huang-TOG:2009}. 
Overall, good results were demonstrated by these works though their applicability scope was limited by the smoothness assumption of the underlying surface, which is rarely matched by data acquired with real scanners. 
To overcome such limitation, in~\cite{Huang-TOG:2013} Huang~\etal~proposed an edge-aware point set resampling solution that first resamples away from edges, then progressively approaches edges and corners. One limitation of this method is the dependence of the quality of the results from the normals accuracy at the points, and the need for a careful tuning of the parameters. 
A point representation method based on volumetric voxelization was introduced by Wu~\etal~\cite{Wu_2015_CVPR}. As a preliminary operation, they proposed to fuse consolidation and completion in one coherent step. However, the goal of this operation was on filling large holes, so that global smoothness is not enforced, making the method sensitive to large noise. 
All these methods are not driven by the data, rather they strongly rely on some priors. 

\smallskip

\textbf{Deep-learning based methods}. 
Only recently, methods have adopted deep architectures to directly learn from point sets. This was mainly due to the inherent difficulty of such data, where points are unordered and do not follow any regular-grid structure in their spatial arrangement. To circumvent such difficulty, some methods converted point clouds to other 3D representations, based on graphs~\cite{Bruna_2014, Masci_2015_ICCV_Workshops} or volumetric grids~\cite{Dai_2017_CVPR, Maturana:2015, Riegler_2017_CVPR, Wu_2015_CVPR}. 
The PointNet~\cite{Qi_2017_CVPR} and Point Net++~\cite{pointnet++} were the first successful attempts to directly process point clouds for classification and segmentation purposes using a hierarchical feature learning architecture that captures both local and global geometry contexts. 
Other networks that were proposed for high-level analysis of point clouds focusing on global or mid-level attributes of point clouds include~\cite{Hua_2018_CVPR, Klokov_2017_ICCV, NEURIPS2018_f5f8590c, Qi_2018_CVPR, Wang_2018_CVPR}. Local shape properties, like normal and curvature in point clouds, were estimated by the network proposed in~\cite{Guerrero-CGF:2018}. 
Interesting network architectures were also proposed for 3D reconstruction from 2D images~\cite{Fan_2017_CVPR, Groueix_2018_CVPR, Lin_Kong_Lucey_2018}. 
For example, Fan~\etal~\cite{Fan_2017_CVPR} addressed the problem of 3D reconstruction from a single image, generating a straight-forward form of output–point cloud coordinates. The 4D extension of the resulting Point Set Generation Network (PSGN-4D) was used in several studies as a baseline for comparison. 

One of the first work aiming to perform point cloud upsampling was proposed by Yu~\etal~\cite{Yu_2018_CVPR}. They introduced the PU-Net that learns per point features at multiple scales, and expands the set of points using a Multi-layer Perceptron (MLP) with multiple branches. However, to learn multi-layer features the input point sets are downsampled, thus potentially causing a loss of resolution. 
In~\cite{Yu_2018_ECCV}, the same authors proposed an edge-aware network for point set consolidation (EC-Net) that uses a specific loss to encourage learning to consolidate points for edges. On the negative side, a very expensive edge-notation is needed for training the EC-Net. 
In the work of Yifan~\etal~\cite{Yifan_2019_CVPR}, a progressive network (3PU) was proposed that duplicates the input point patches over multiple steps. The progressive architecture of 3PU makes its training computationally expensive. More data are also required to supervise the middle stage outputs of the network. 
A Generative Adversarial Network designed to learn upsampled point distributions (PU-GAN) was proposed by Li~\etal~\cite{Li_2019_ICCV}, with the main performance improvement obtained by the discriminator. 
Qian~\etal~\cite{Qian-eccv:2020} proposed to upsample points by learning the first and second fundamental forms of the local geometry. However, their PUGeo-Net needs additional supervision in the form of normals.
The PU-GCN proposed by Qian~\etal~\cite{Qian_2021_CVPR} performed upsampling by leveraging on an Inception based module to extract multi-scale information, and using a GCN-based upsampling module to capture local point information. This has the main advantage of not needing for additional annotations, like edges, normals, point clouds at intermediate resolutions, \etc, while also avoiding the use of a sophisticated discriminator.

Recently more and more works have been shifting the attention towards \textit{4D reconstruction}, where a sequence of 3D objects is reconstructed from time-varying point clouds given as inputs~\cite{4D_Leroy_2017_ICCV, 4D_Mustafa_2016_CVPR}.

In the Occupancy Network (ONet) proposed by Mescheder~\etal~\cite{mescheder2019occupancy}, a 3D object was described using a continuous indicator function that indicates which sub-sets of the 3D space the object occupies, and an iso-surface retrieved by employing the Marching Cube algorithm. 
Tang~\etal~\cite{Tang2021LearningPD} learned a temporal evolution of the 3D human shape through spatially continuous transformation functions among cross-frame occupancy fields. To this end, they established, in parallel, the dense correspondence between predicted occupancy fields at different time steps via explicitly learning continuous displacement vector fields from spatio-temporal shape representations. 
Niemeyer~\etal~\cite{Niemeyer_2019_ICCV} introduced a learning-based framework for object reconstruction directly from 4D data without predefined templates. The proposed OFlow method calculates the integral of a motion field of 3D points in a 3D point cloud specified in space and time to implicitly represent trajectories of all the points in dense correspondences between occupancy fields.
Vu~\etal~\cite{tavu2022rfnet4d} proposed a network architecture, called RFNet-4D, that jointly reconstructs objects and their motion flows from 4D point clouds. It is shown that jointly learning spatial and temporal features from a sequence of point clouds can leverage individual tasks, leading to improved overall performance. To this end, a temporal vector field learning module using unsupervised learning approach for flow estimation was designed that, in turn, leveraged by 
supervised learning of spatial structures for object reconstruction.
Jiang~\etal~\cite{Jiang_2021_CVPR} introduced a compositional representation that disentangles shape, initial state, and motion for a 3D object that deforms over a temporal interval. Each component is represented by a latent code via a trained encoder. A neural Ordinary Differential Equation (ODE) is used to model the motion: it is trained to update the initial state conditioned on the learned motion code, while a decoder takes the shape code and the updated state code to reconstruct the 3D model at each time stamp. An Identity Exchange Training (IET) strategy is also proposed to encourage the network to learn decoupling each component. 

\smallskip

With respect to the above solutions, our approach is characterized by a specific design that combines two Graph Convolutional Networks (GCNs) to work in an adversarial setting (GAN). The resulting architecture proved to be flexible in the number of frames used as inputs and conjugated effective reconstructions with inference times that are compatible with online execution. 

\section{Problem Statement}\label{sec:problem}
We consider a sequence of point clouds in the 3D space. Each point cloud can be regarded as a frame of a 4D video at time $t$.
In the following, we consider $n$ point cloud frames \textit{fused} together forming a time varying point cloud as an unordered lists of ${x,y,z,t}$ points. Our task is to \textit{upscale}, a term borrowed from the 2D image super-resolution domain, each of the point cloud (frame) of the input sequence $F_t$ and get a more detailed one by leveraging the information of the previous $n-1$ low-resolution point cloud frames (\ie, $F_{t-1}, \dots, F_{t-n+1}$).

More in detail, given a buffer composed of $n$ previous frames, the input point cloud $P_i$ is defined as:
\begin{equation}
P_i = \{p_{-n+1}, p_{-n+2}, ... , p_{0}\} ; P_i \in \mathbb{R}^{\{x,y,z,t\ \times L \times n\}}, 
\end{equation}

\noindent
where each low-resolution point cloud $p_i$ is composed of $L$ points:
\begin{equation}
p_i \in \mathbb{R}^{\{x,y,z,t\ \times L\}}.
\end{equation}

We are interested in learning a map $f(P_i,\theta)$ from $P_i$ $\rightarrow$ $P_T$, where $P_T$ is the target point cloud and it represents the zeroth frame upscaled to have $H = S \times L \times n$ points, with $S$ being the \textit{scale factor}.
\begin{equation}
    P_T \in \mathbb{R}^{\{x,y,z\ \times H\}}.
\end{equation}

\section{Proposed Method}
Our proposed method makes use of message passing Graph Networks, different neighbourhood sampling techniques and Generative Adversarial training.

We employed a Graph Neural Network for this task. More in detail, our architecture has been developed starting from~\cite{pointnet++}. The employed architecture works on unordered lists of ${x,y,z,t}$ points, representing the last $n$ frames \textit{fused} together, using two Graph Convolutional Neural Networks (GCNs from here on) in an adversarial setting.
The discriminator is based on~\cite{pointnet++}, while the generator improves on the architecture proposed in~\cite{pointnet++}. 
In particular, we used different neighbors sampling techniques that were developed with the intent of collecting, for each point, features contemporaneously of its immediate neighborhood and also from furthest vertices of the whole point cloud without making the computation too expensive.

The fully convolutional nature of our generator network allows us to potentially train and test at different input and output resolutions. 

\subsection{Edge Convolution and GAT}
The basic module composing our generator network is made of the combination of Edge Convolution~\cite{wang2019dynamic} and Graph Attention Networks (GAT)~\cite{velivckovic2017graph}. The Edge Convolution allows us to perform message passing over a dynamic graph in which the edges are updated as the point cloud changes. The GAT side is used to perform an attentional aggregation over the features collected from the dynamic local neighbourhood, this in contrast with much more common choices for aggregation such as $max$ or $average$. We refer to this combination module as \textit{Edge Convolution with Attention}.

\subsection{Parallel Double Sampling (PDS) module}\label{subsec:samgen}
\begin{figure}[!ht]
\centering
\includegraphics[width=\linewidth]{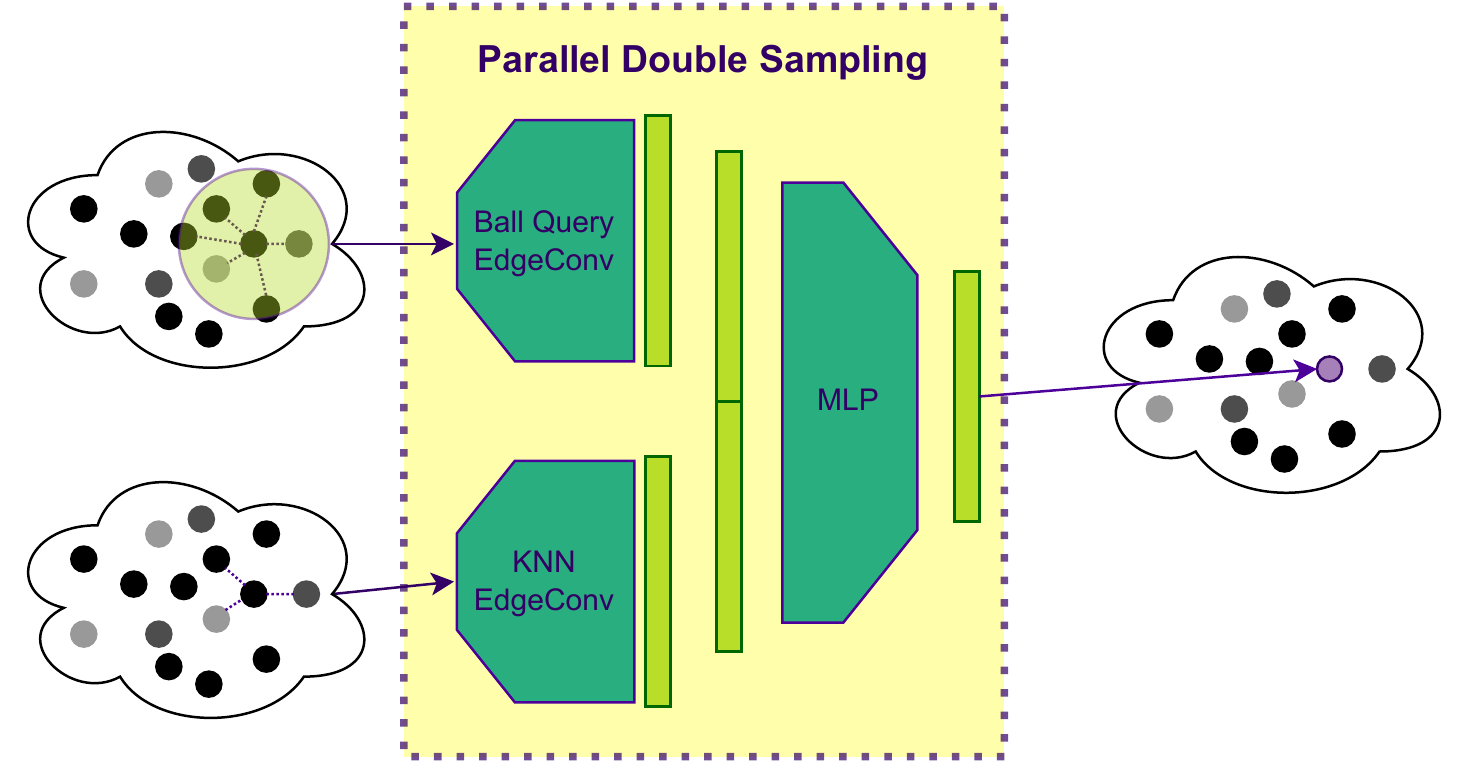}
\caption{Schematic representation of the proposed Parallel Double Sampling (PDS) module.}
\label{fig:samgen}
\end{figure}

The core of the generator side of the architecture is the Parallel Double Sampling (PDS) module that performs two different graph convolutions using two different sets of sampled points. A simplified illustration of this module is presented in Figure~\ref{fig:samgen}.
For each point, two sets of operations are performed in a parallel fashion. 
The first set, is a pipeline composed of:
\begin{itemize}
    \item \textbf{Radius filtering}: For each vertex, a filtering step leaves as neighbors, with the capability of passing messages, only those vertices that belong to a sphere of radius $r$, centered on the vertex.
    
    \item \textbf{Furthest Point Subsampling}: We use the Furthest Point Subsampling (FPS) algorithm in~\cite{pointnet++} in order to sample temporarily, a fraction $s$ of the original points that are the farthest away, inside the radius, from a starting point.

    \item \textbf{Convolution}: Graph convolution is applied over the remaining vertices, \textbf{independently of their number}, and their features are aggregated.
\end{itemize}

\noindent
The second set of operations, performed in parallel to the first one, is composed of:
\begin{itemize}
    \item \textbf{K-NN}: A fixed number of $k$ closest vertices is selected as neighbors.
    \item \textbf{Convolution}: Graph convolution is applied over the vertices, and their aggregated features.
\end{itemize}

\noindent
Finally, the two sets of features are concatenated and fed to a linear layer that maps $2 \times Channels_{in} \rightarrow Channels_{out}$.

\subsection{Our Architecture}
The developed architecture is composed of two Graph Convolutional Networks (GCNs) working in an adversarial setting (GAN)~\cite{NIPS2014_5ca3e9b1}.
It is illustrated in the bottom of Figure~\ref{fig:gcn-arch}. Basically, the point cloud given as input is processed as a graph using message passing based convolution. 

\subsection{Discriminator}
The discriminator is inspired by the PointNet++ architecture~\cite{pointnet++}, since it also targets a classification task. We use the same structure that progressively reduces the number of points using \textit{max-pooling} operations and finally a sequence of linear layers before the output as shown in the bottom part of Figure~\ref{fig:gcn-arch}.

\subsection{Generator}
\begin{figure}[!th]
\centering
\includegraphics[width=\linewidth]{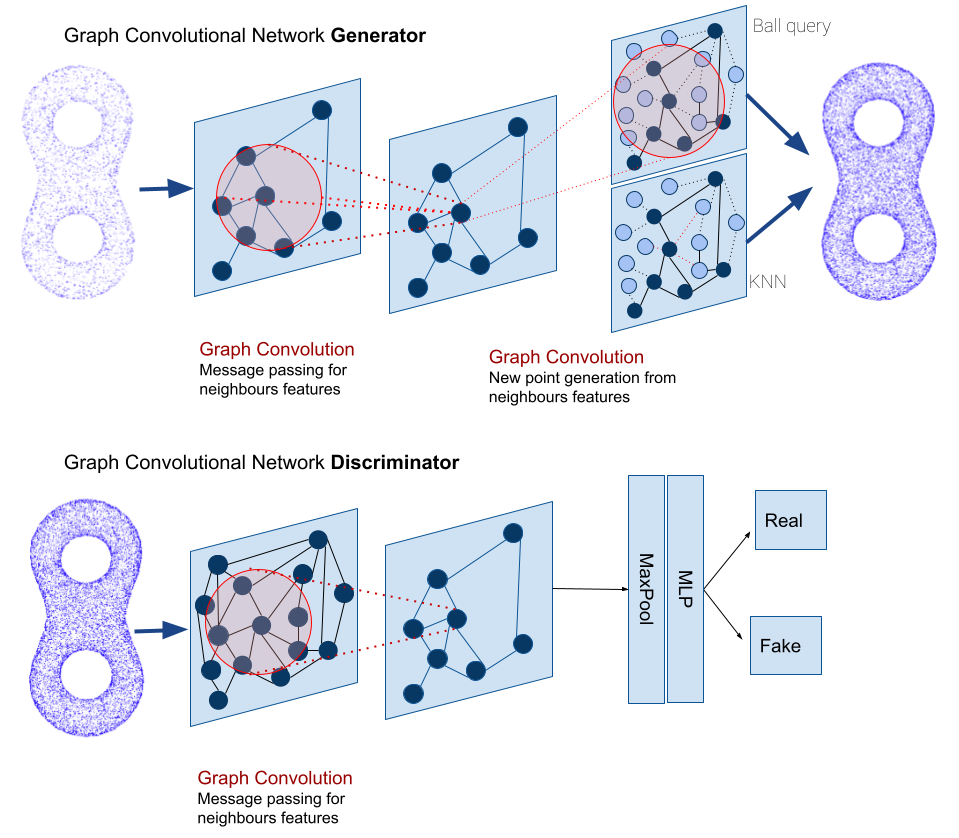}
\caption{Schematic representation of the proposed GCN architecture. \textit{Top}: Generator architecture;
\textit{Bottom}: Discriminator architecture.}
\label{fig:gcn-arch}
\end{figure}

The generator side of the model is instead built as an initial sequence of Edge Convolution with Attention modules followed by our Parallel Double Sampling (PDS) module. It is also inspired by the PointNet++ architecture~\cite{pointnet++} but undergoing major changes as detailed in Section~\ref{subsec:samgen}. 
In the top part of Figure~\ref{fig:gcn-arch} a simplified visualization of the PDS generator is presented. 
The generator is composed of multiple Graph Convolutions with Attention followed by a single PDS. The intuition behind this choice is to collect various features for each node, using different neighborhood sampling techniques. Once the original node has been enriched with the local features, the PDS will use them to generate multiple new vertices according to the scale factor. Finally this new vertices position is summed with the closest one that originated it, in a sort of residual fashion (see Figure~\ref{fig:teaser}).

The generator loss $L_g$ is composed of an adversarial component $L_{adv}$ coming from the Discriminator,a full reference reconstruction loss $L_{rec}$ computed as the Chamfer Distance between the restored point cloud and the original one and an additional Density Loss $L_{D}$.
We used the LSGAN from~\cite{Mao2016LeastSG} loss for our training, which assumes the form:
\begin{equation}
L_{adv} = \min_{G}L\left(G\right) = \frac{1}{2}\mathbb{E}_{\mathbf{z} \sim p_{\mathbf{z}}\left(\mathbf{z}\right)}\left[\left(D\left(G\left(\mathbf{z}\right)\right) - c\right)^{2}\right],
\end{equation}

\noindent
for the generator, and:
\begin{eqnarray}
    \min_{D}L\left(D\right) = \frac{1}{2}\mathbb{E}_{\mathbf{x} \sim p_{data}\left(\mathbf{x}\right)}\left[\left(D\left(\mathbf{x}\right) - b\right)^{2}\right] + \\ 
    + \frac{1}{2}\mathbb{E}_{\mathbf{z}\sim p_{\mathbf{z}}\left(\mathbf{z}\right)}\left[\left(D\left(G\left(\mathbf{z}\right)\right) - a\right)^{2}\right], 
\end{eqnarray}

\noindent
for the discriminator.

\subsection{Loss functions}
The model is trained end-to-end using multiple losses. Beside the adversarial component $L_{adv}$, we also compute the point-to-set distance (Chamfer distance) $L_{rec}$ between the reconstructed point cloud and the target one and, similarly to~\cite{wu2021density}, we take into account the \textit{neighbourhood} of each point. That is, for each reconstructed point $p_{r} \in P_r$, we find the closest point $p_{t} \in P_t$ in the target point cloud, and compute both the distance between them and the difference in terms of local neighbors:
\begin{equation}
L_{CD}(P_{r},P_{t}) = \sum_{r \in P_{r}}\operatorname*{min}_{t \in P_{t}} ||r-t||^2_2 + \sum_{t \in P_{t}} \operatorname*{min}_{r \in P_{r}} ||r-t||^2_2 .
\label{eq:chamfer}
\end{equation}

We define a vertex $p$ \textit{neighbourhood} density $D(p)$ as the normalized sum of its neighbours in a given radius:
\begin{equation}
D({p \in P}) = \frac{1}{N_{max}} \sum_{n \in Ball_{p}} 1 , 
\end{equation}

\begin{equation}
L_{D}(P_{r},P_{t})= \sum_{r \in P_{r}}\operatorname*{min}_{t \in P_{t}} ||D(r)-D(t)||^2_2 + \sum_{t \in P_{t}} \operatorname*{min}_{r \in P_{r}} ||D(r)-D(t)||^2_2 .
\end{equation}

The generator final loss is therefore given by:
\begin{equation}
L_{rec} = \lambda_1 L_{CD}+ \lambda_2 L_{D} + \lambda_3 L_{Adv} ,
\end{equation}

\noindent
where values for $\lambda_i$ have been empirically determined ($\lambda_1=1.0 , \lambda_2 = 0.5 , \lambda_3=0.1$).

\section{Experiments}\label{sec:results}
The proposed solution for point clouds upscaling has been evaluated in a comprehensive set of experiments. both qualitative (Section~\ref{sec:qualitative}) and quantitative (Section~\ref{sec:quantitative}). An ablation study aiming to evidence the relevance of different components of our architecture is also reported in Section~\ref{sec:ablation}. 

\subsection{Implementation details}
Our model is implemented in PyTorch, using the PyTorch Geometric (PyG) library~\cite{fey2019fast}. This library was built upon PyTorch and is specifically designed for Graph Neural Networks (GNNs).
The two networks are implemented as two Message Passing Networks put in an adversarial setting. Both the Discriminator and the Generator are optimized with Adam, using the standard learning rate $lr = 1e^{-4}$ and betas $\beta_1 = 0.9, \beta_2 = 0.999$, using a linear decaying scheduler that drops the learning rate to 1/10th every 10 epochs.
Other hyperparameters, such as the radii for the \textit{Ball Query} for the FPS sampling ($r_{small}=0.06$, $r_{large}=0.1$) and the number of neighbours for the \textit{KNN} sampling ($n_{neighbours}=9$) were empirically determined trough grid search.
\subsubsection{Augmentation}
The training data is augmented using different operations. Each sequence of input point clouds and its relative ground truth point cloud is randomly flipped along any of its axes, per point random noise is added, and finally a random scale along any axes is applied in a range between $[0.9,1.1]$.
As a form of augmentation, we also exploit the fully convolutional nature of the generator architecture; similar to the case of 2D image super-resolution, where patches of the target high resolution image are used in the training, we randomly feed a 3D slice of the video instead of the full body.
A final form of augmentation used during training is a simple time inversion inside the sequence.

\newcommand{\wdl}{0.24\linewidth}

\begin{figure*}[t]
\begin{center}
\begin{tabular}{c|c}
\includegraphics[width=\wdl]{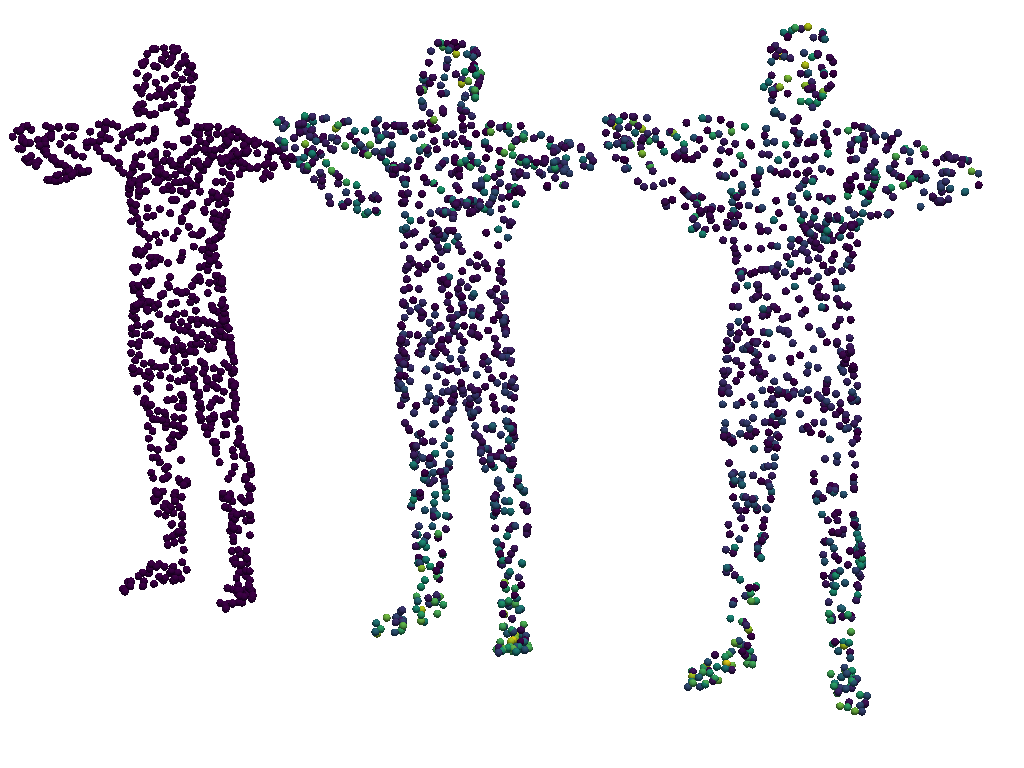}&
\includegraphics[width=\wdl]{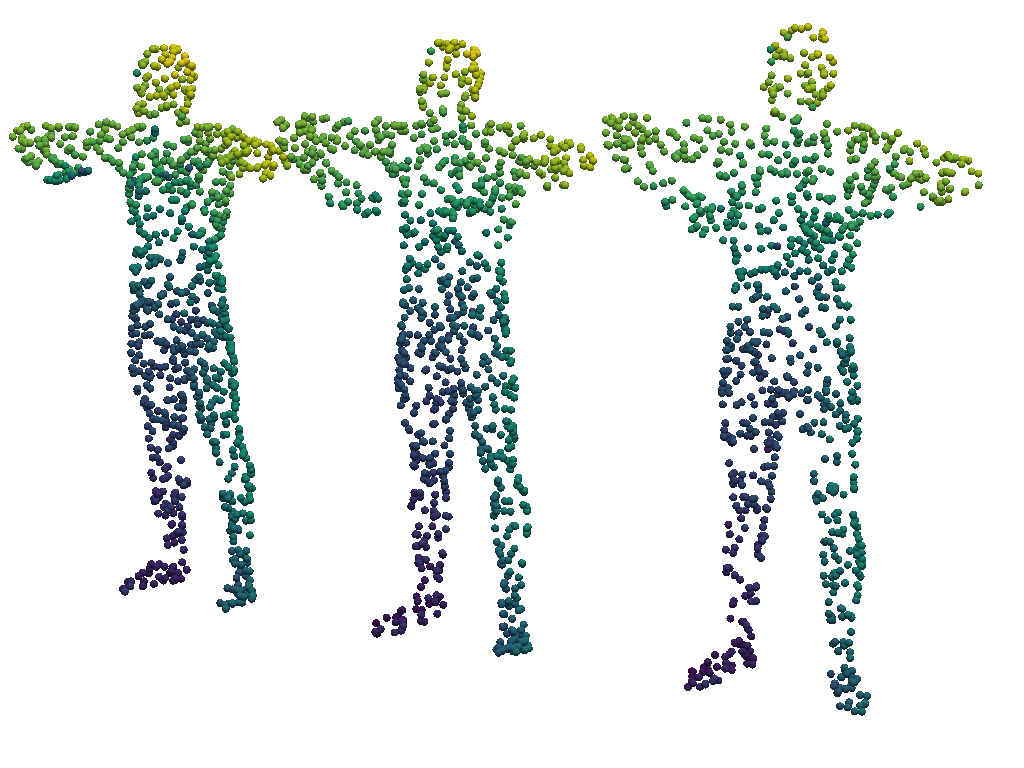}
\includegraphics[width=\wdl]{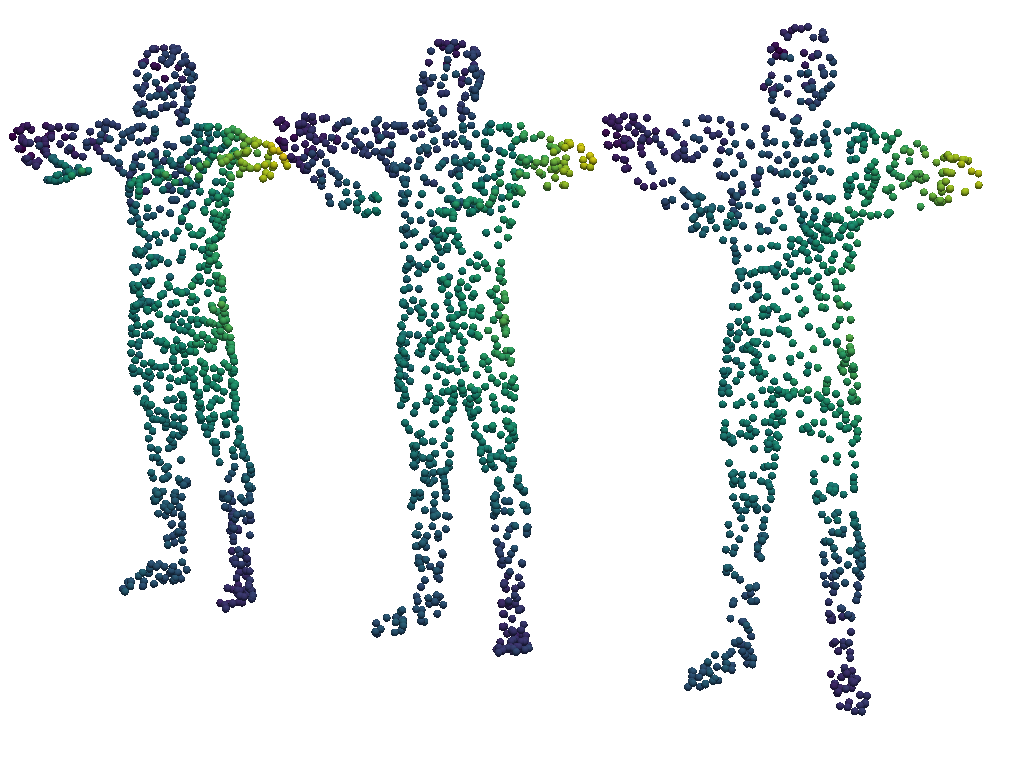}
\includegraphics[width=\wdl]{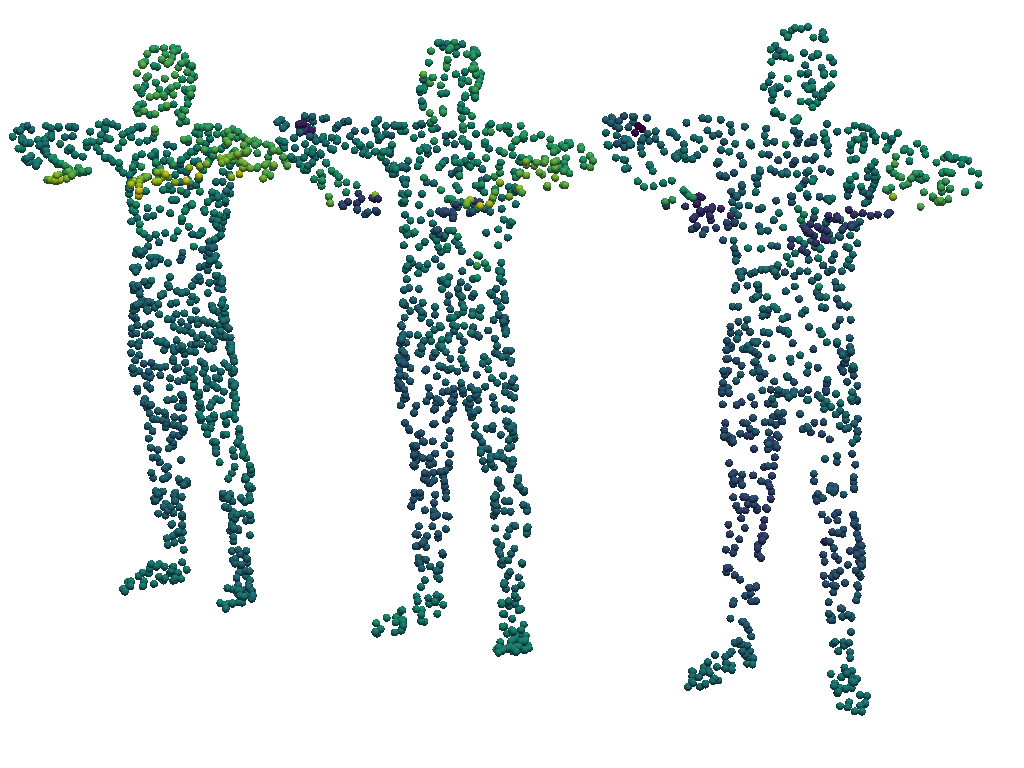}
\end{tabular}
\\
\begin{tabular}{c|c}
\includegraphics[width=\wdl]{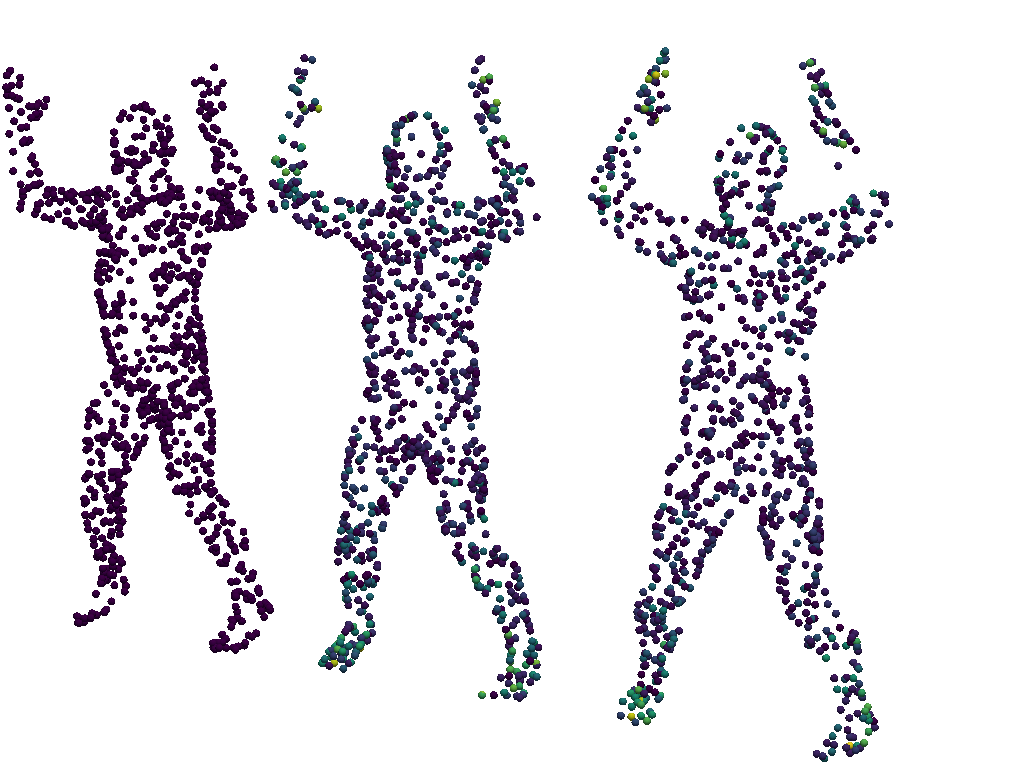}&
\includegraphics[width=\wdl]{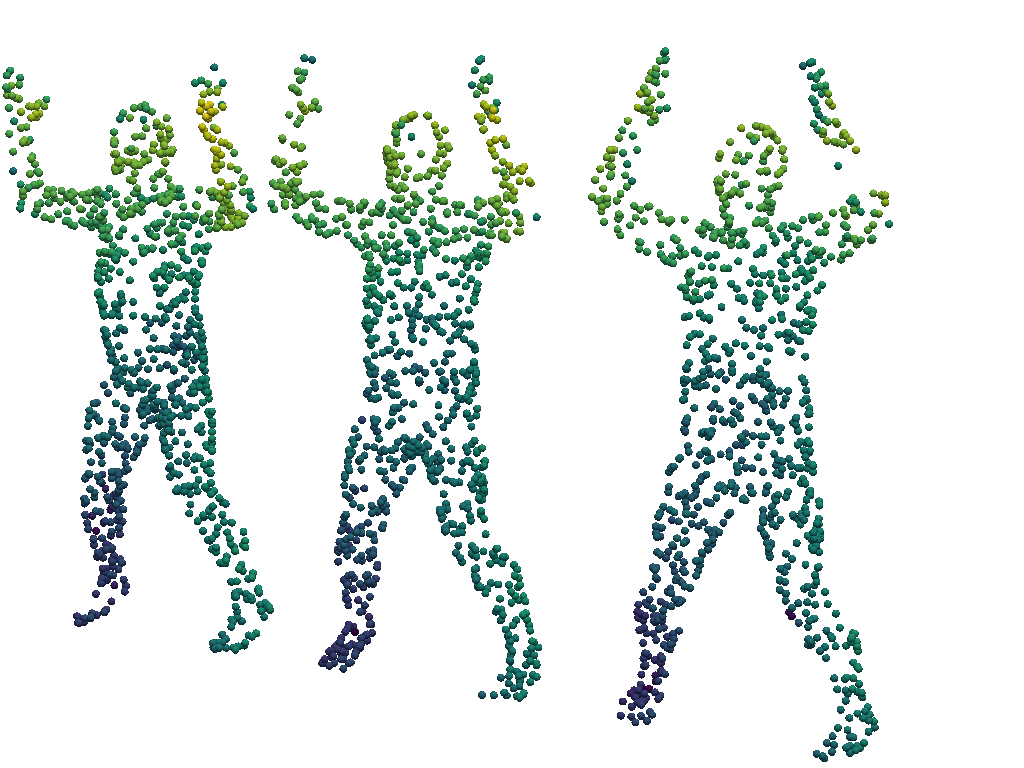}
\includegraphics[width=\wdl]{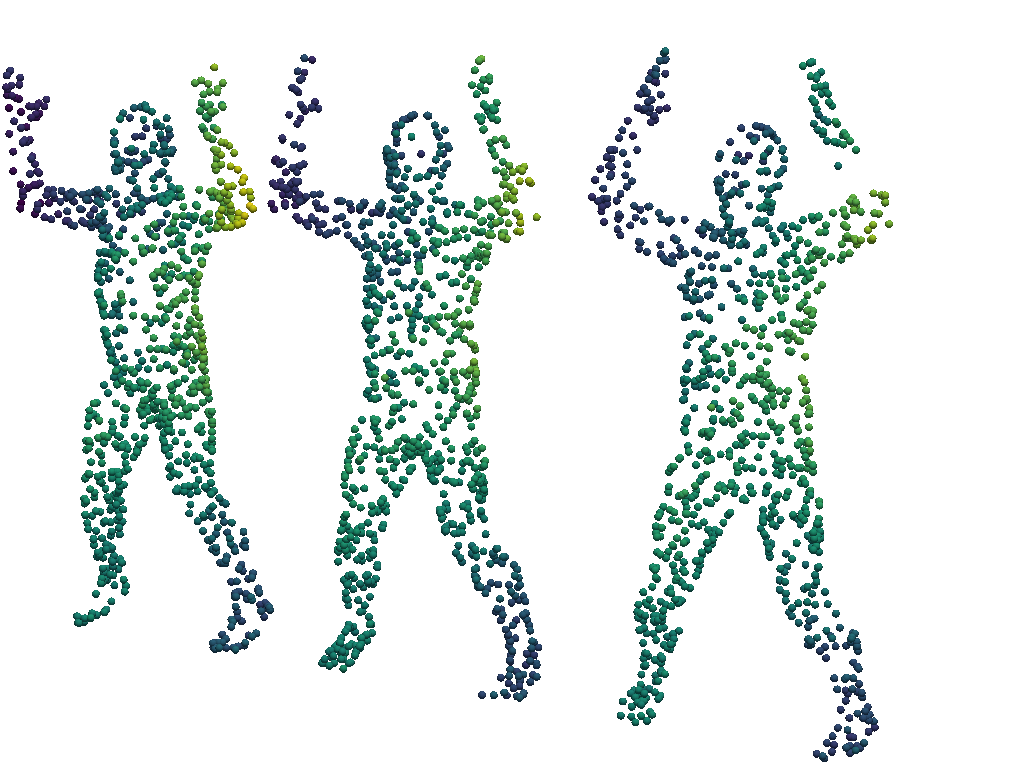}
\includegraphics[width=\wdl]{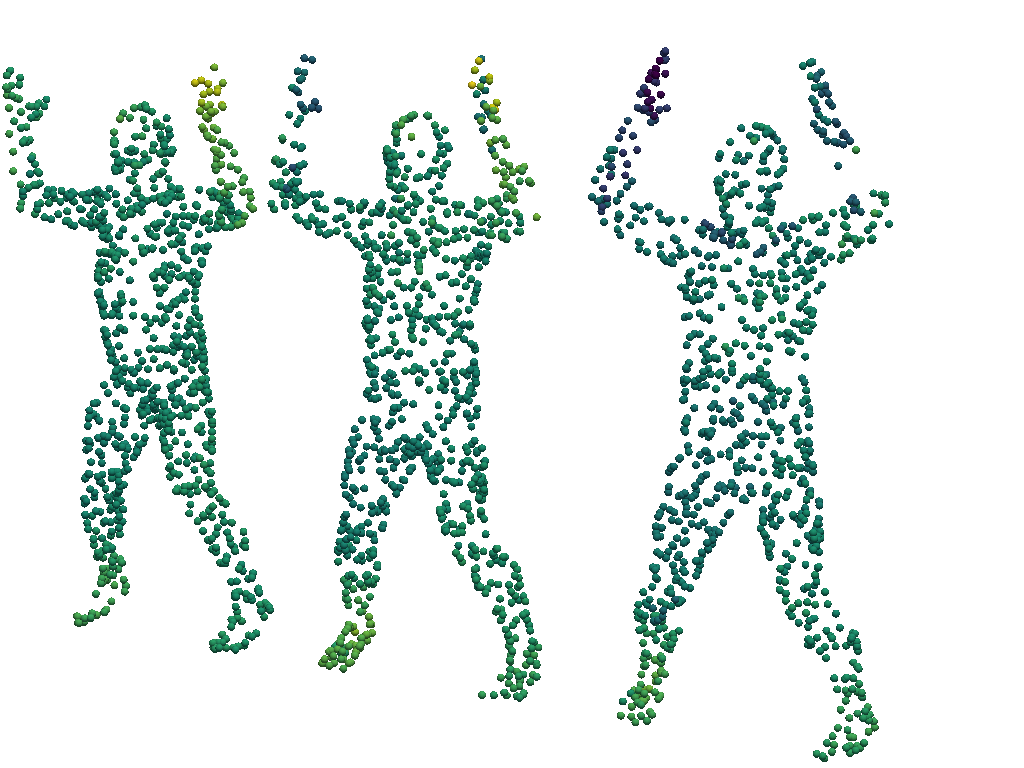}\\
\end{tabular}
\end{center}
\caption{The top and bottom row show: (\textit{left}) the point clouds of a three-frame input sequence with movements. Colors indicate the movement of a point with respect to the previous frame; (\textit{right}) different features obtained from subsequent edge graph convolutional layers of the proposed architecture as a response to the three-frame sequence shown on the left. It can be noted the response of layers seems to pass from spatial to temporal details.}
\label{fig:detland}
\end{figure*}

\subsection{Dataset}
To evaluate our proposed solution, we used the Dynamic FAUST (D-FAUST) dataset~\cite{dfaust:CVPR:2017}. It contains animated meshes for 129 sequences of 10 human subjects (5 females and 5 males) with various motions such as ``shake hips'', ``punching'', running on ``spot'', or ``one leg jump''. 

In order to compare with other methods, we used the train/test split proposed in~\cite{Niemeyer_2019_ICCV}. For each sequence, at training time, we randomly pick an index and then subsample the following frames according to the model's frame rate. We trained multiple models at different frame rates.

We also followed the evaluation setup used in~\cite{Niemeyer_2019_ICCV}. Specifically, for each evaluation, we carried out two case studies: seen individuals but unseen motions (\ie, test subjects were included in the training data but their motions were not given in the training set); and unseen individuals but seen motions (\ie, test subjects were found only in the test data but their motions were seen in the training set).

\subsection{Qualitative results}\label{sec:qualitative}
\begin{figure}[ht!]
\centering
\includegraphics[width=\linewidth]{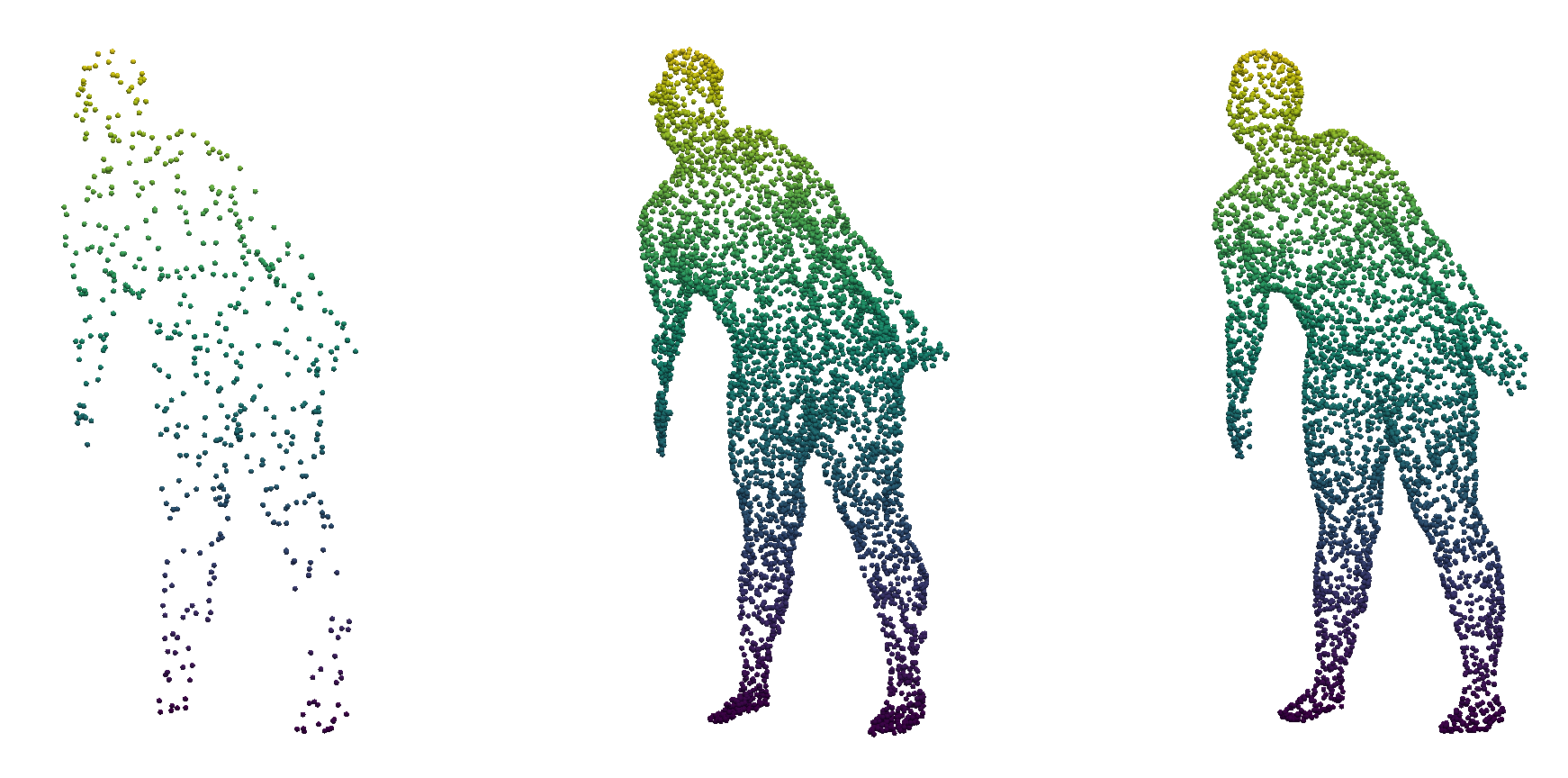}
\caption{\textit{Left:} Sample of a single frame from an input low resolution point cloud with $\sim$512 vertices, \textit{Center:} reconstruction obtained with our proposed solution; \textit{Right:} Ground truth point cloud.}
\label{fig:qualitative01}
\end{figure}

\begin{figure}[ht!]
\centering
\includegraphics[width=\linewidth]{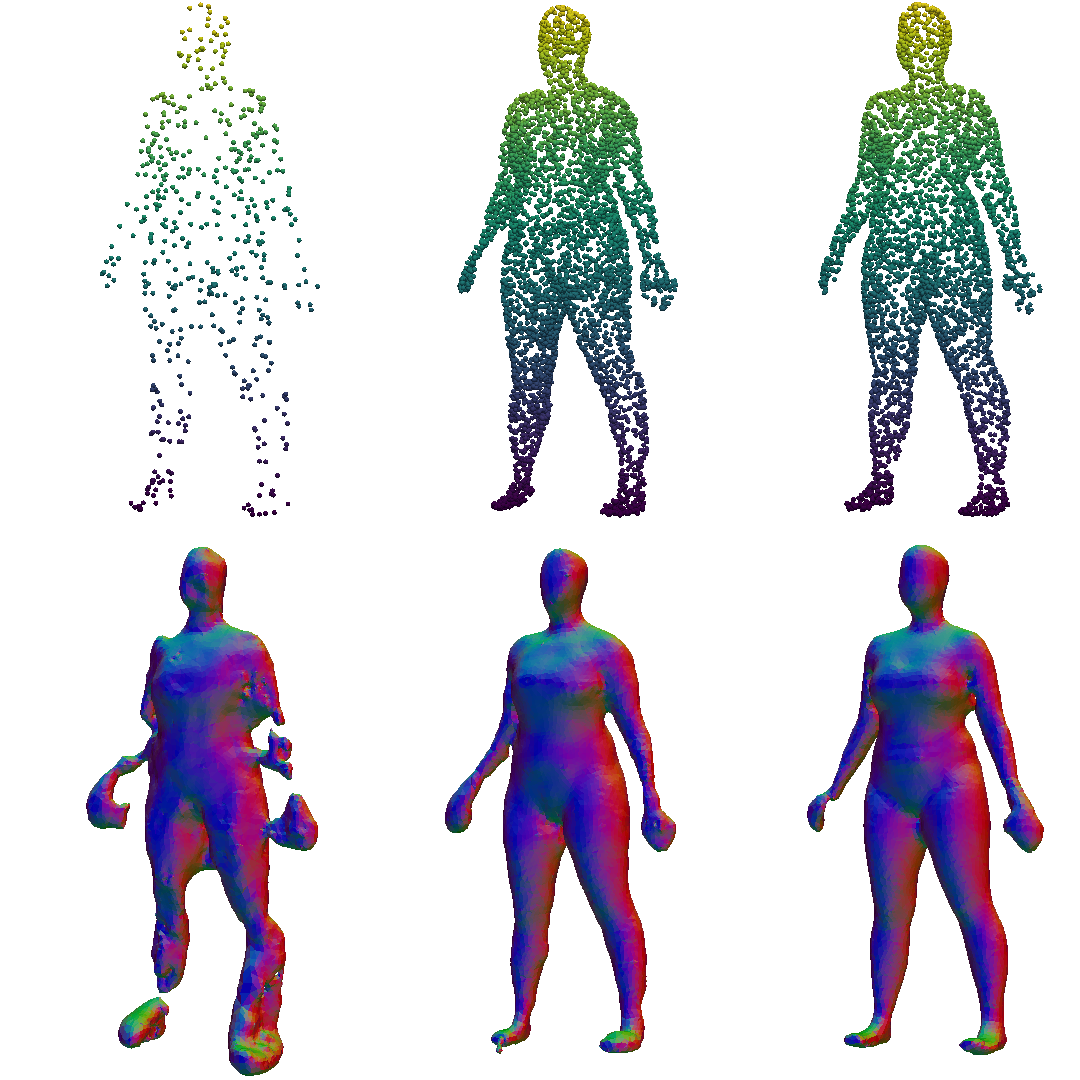}
\caption{\textit{Top:} Point cloud visualization. \textit{Bottom:} Mesh reconstruction using Poisson surface reconstruction from~\cite{poisson_rec}.
\textit{Left:}  Sample obtained from a single frame of an input low-resolution point cloud with 1024 vertices; \textit{Center:} Model reconstruction using our proposed approach; \textit{Right:} Ground truth point cloud.}
\label{fig:qualitative_mesh}
\end{figure}

Some qualitative results of the proposed upscaling are given in Figures~\ref{fig:qualitative01} and~\ref{fig:qualitative_mesh}. 
In Figure~\ref{fig:qualitative01}, the input low-resolution frame, our reconstruction point cloud and the ground truth are given from left ro right. 
A second example is shown in Figure~\ref{fig:qualitative_mesh}, where the input frame, our reconstruction and the ground truth are compared both in terms of point clouds (top) and in terms of mesh reconstruction using the Poisson algorithm (bottom). 
Additional qualitative results are given as videos in the supplementary material.

To give some insights on the behaviour of the network layers, we inspected the response of the various convolutional layers given an input point cloud, and visualized them. 
As an example, on the left of Figure~\ref{fig:detland}, three frames of an input point cloud are shown (the frames are taken at three consecutive times, $t_0$, $t_0+1$ and $t_0+2$). Points in the clouds are colored to highlight their movement with respect to the previous frame. 
On the right of Figure~\ref{fig:detland}, instead, the response features of different layers are visualized (the depth of the layers increases from left to right). It is interesting to note as, similarly to CNNs, depth correlates to complexity: The first convolutional layers seem to have strong response for large physical parts of the human subject, while the latter stages take into consideration time and movement. 

\subsection{Quantitative results}\label{sec:quantitative}
\subsubsection{Evaluation Metrics}
To measure the reconstruction quality we applied the standard Chamfer Distance (CD), a point-to-set metric since following the same protocol as reported in~\cite{tavu2022rfnet4d} that uses the CD as reconstruction metric for measuring the dissimilarity between a point and a point set.

\subsubsection{Compared Methods}
We compared our approach with respect to six state-of-the-art solutions in the literature for 4D reconstruction from point cloud sequences, namely, PSGN 4D, ONet 4D, OFlow, LPDC, 4DCR, and RFNet-4D. 
The PSGN 4D extends the PSGN approach~\cite{Fan_2017_CVPR} to predict a 4D point cloud, \ie, the point cloud trajectory instead of a single point set. 
The ONet 4D network is an extension of ONet~\cite{mescheder2019occupancy} to define the occupancy field in the spatio-temporal domain by predicting occupancy values for points sample in space and time. 
The OFlow network~\cite{Niemeyer_2019_ICCV} assigns each 4D point an occupancy value and a motion velocity vector and relies on the differential equation to calculate the trajectory. 
The LPDC~\cite{Tang2021LearningPD} learned a temporal evolution of the 3D human shape through spatially continuous transformation functions among cross-frame occupancy fields. 
The 4DCR solution~\cite{Jiang_2021_CVPR} used a compositional representation that disentangles shape, initial state, and motion for a 3D object that deforms over a temporal interval. 
Finally, RFNet-4D~\cite{tavu2022rfnet4d} jointly reconstructs objects and their motion flows from 4D point clouds. 

\subsubsection{Results}
Tables~\ref{tab:results-uism} and~\ref{tab:results-sium} report results for our solution and for the other methods as given in~\cite{tavu2022rfnet4d}. 
For our method (last line in the tables) we used 3 frames for upscaling at 60fps with a scale factor of $\times4$ starting from low-resolution point clouds composed of 1024 vertices.
For the \textit{unseen individual and seen motion} protocol in Table~\ref{tab:results-uism}, our approach achieves the second best score. 
From Table~\ref{tab:results-sium}, it can be observed that our method reached a reconstruction error of similar magnitude with respect to the two best performing methods, \ie, RFNet-4D and LPDC. 
It is worth noting that RFNet-4D obtained the reported error using a larger number of input frames (\ie, 17 against 3 to 8 as used in our tests). It was not possible to test the RFNet-4D with our setting because the code was not publicly available. 

\begin{table}[!ht]
\begin{center}
\begin{tabular}{l|c}
\hline
Method & Chamfer Distance x $10^{-3}\downarrow$ \\
\hline
PSGN-4D~\cite{Fan_2017_CVPR} & 0.6877 \\
ONet-4D~\cite{mescheder2019occupancy} & 0.7007 \\
OFlow~\cite{Niemeyer_2019_ICCV} & 0.2741 \\
4DCR~\cite{Jiang_2021_CVPR} & 0.2220 \\
LPDC~\cite{Tang2021LearningPD} & 0.2188 \\
RFNet-4D~\cite{tavu2022rfnet4d} & \textbf{0.1594} \\
\hline
Ours & \underline{0.1758}\\
\hline
\end{tabular}
\end{center}
\caption{Reconstruction accuracy for the \textit{unseen individuals and seen motions} protocol. We report the Chamfer distance (lower is better). Results for the best and second best performing methods are given in bold and underlined, respectively. Our approach scored the second best accuracy. 
\label{tab:results-uism}}
\end{table}

\begin{table}[!ht]
\begin{center}
\begin{tabular}{l|c}
\hline
Method & Chamfer Distance x $10^{-3}\downarrow$ \\
\hline
PSGN-4D~\cite{Fan_2017_CVPR} & 0.6189 \\
ONet-4D~\cite{mescheder2019occupancy} & 0.5921 \\
OFlow~\cite{Niemeyer_2019_ICCV} & 0.1773 \\
4DCR~\cite{Jiang_2021_CVPR} & 0.1667 \\
LPDC~\cite{Tang2021LearningPD} & \underline{0.1526} \\
RFNet-4D~\cite{tavu2022rfnet4d} & \textbf{0.1504} \\
\hline
Ours & 0.1638\\
\hline
\end{tabular}
\end{center}
\caption{Reconstruction accuracy for the \textit{seen individuals and unseen motions} protocol. We report the Chamfer distance (lower is better). Results for the best and second best performing methods are given in bold and underlined, respectively. Our approach results in the third best performance.  
\label{tab:results-sium}}
\end{table}

In Table~\ref{tab:time-results}, we report the inference time, in seconds, for various different configurations of our model. All the measurements correspond to experiments executed on an Nvidia 2080Ti GPU. The values reported in the table evidence that our approach can open the way to real-time upscaling.
As reported in~\cite{tavu2022rfnet4d}, their method used 17 inout frames to reconstruct an output frame, while our range of frames is between 3 (for models using larger input point clouds) and 8 (for smaller inputs) due to memory constraint {at training time}.

\begin{table}[!ht]
\begin{center}
\begin{tabular}{l|c|c|c}
\hline
Method & Input size & Upscale $\times$ & Inference time (s) $\downarrow$ \\
\hline
Ours & 1024 & 3 & 0.103\\
Ours & 1024 & 2 & 0.089\\
Ours & 512 & 4 & 0.046\\
Ours & 512 & 2 & 0.039\\
Ours & 256 & 8 & 0.034\\
Ours & 256 & 4 & 0.030\\
\hline
Oflow\cite{mescheder2019occupancy} & - & - & 0.95\\
LDPC\cite{Tang2021LearningPD} & - & - & 0.44\\
RfNet4d\cite{tavu2022rfnet4d} & - & - & 0.24\\
\hline
\end{tabular}
\end{center}
\caption{Inference time for different configurations of our model using a three-frames buffer. Every test was performed on an Nvidia2080Ti. For the other models it must be noted that they used a 17 frame input sequence to output a frame.
\label{tab:time-results}}
\end{table}

\subsection{Ablation Studies}\label{sec:ablation}
In this section, we present ablation studies to verify different aspects in the design of our model. 
In particular, we verify each of the introduced components in our architecture for 4D point clouds reconstruction by comparing the percentage decrease of the model when some particular features are removed.

We performed a first set of experiments by using a stream of input point clouds at 60fps and with 256 points per frame; on this stream, we performed upscaling from subsets of consecutive 3 frames, using an upscale factor of $\times$2.
From Table~\ref{tab:ablation-256}, we can notice that by removing individual components of our architecture, the performance of the model significantly and consistently decreases.
In particular, we removed the attention aggregation module and we substituted it with a more common \textit{mean} aggregation.
We also ablated the impact of the Density Loss and the adversarial component.

\begin{table}[!ht]
\begin{center}
\begin{tabular}{l|c|c}
\hline
Variant & Chamfer x $10^{-3} \downarrow$ & \% wrt F. Featured\\
\hline
No Attention & 1.193 & +3.11\% \\
No Density Loss & 1.226 & +5.96\% \\
No Adversarial Loss & 1.213 & +4.84\% \\
\hline
Ours Fully Featured &  \underline{1.157} & - \\
\hline
\end{tabular}
\end{center}
\caption{Ablation study for our model using 256 input points, 3 frames, 60fps, and upscale factor $\times$2.
\label{tab:ablation-256}}
\end{table}

In Table~\ref{tab:ablation-512}, we repeated the above ablation experiments using a different setup. In this case, the frame rate is changed to 30fps, the input resolution to 512 points per frame, and we performed upscaling using a factor of $\times$4.

\begin{table}[!ht]
\begin{center}
\begin{tabular}{l|c|c}
\hline
Variant & Chamfer x $10^{-3} \downarrow$ & \% wrt F. Featured\\
\hline
No Attention & 0.5856 & +2.82\%\\
No Density Loss & 0.6433 & +12.95\%\\
No Adversarial Loss & 0.5930 & +4.13\% \\
\hline
Ours Fully Featured &  \underline{0.5695} & - \\
\hline
\end{tabular}
\end{center}
\caption{Ablation study for our model using 512 input points, 3 frames, 30fps, and upscale factor $\times$4.\label{tab:ablation-512}}
\end{table}

Also in this case, ablating the density loss term results into the most significant decrease in the accuracy of the upscaled model. It is also interesting to observe that, while the percentage increment in the Chamfer distance when removing the attention layer and the adversarial loss shows small differences between the two tables, this is not the case for the density loss: removing this term has a much larger impact on the results in Table~\ref{tab:ablation-512} ($\sim+13\%$) than in Table~\ref{tab:ablation-256} ($\sim+6\%$).

\subsubsection{Importance of the temporal information}
A question that arises with the proposed solution is the actual impact of having the time buffer compared to using just the last point cloud as an input. 
To compare these two solutions, we feed our model with the same frame repeated $n$ times. In this way, we keep the comparison fair by not changing the input size and the amount of starting points but only the \textit{information} contained within it. We refer to this setup as \textit{Static Sequence}, whilst we use the term \textit{Dynamic} to refer the proposed procedure that uses $n$ different frames. 
In Table~\ref{tab:ablation-time}, we report some comparative results between the two ways of using the frames in a sequence. 
It can be observed that there is useful information in the time and movement of the cloud. Just like in a 2D video, the same frame repeated $n$ times does not contain the same amount of useful data for reconstruction as $n$ different subsequent frames.

\begin{table}[!ht]
\begin{center}
\begin{tabular}{l|l|l|l|c}
\hline
Sequence & Input size & Frms & $\times$  & Chamfer x $10^{-3} \downarrow$ \\
\hline
\textit{Static} & 256 & 3 & 4 & 2.876\\
\textit{Dynamic} & 256 & 3 & 4 &  \underline{1.109}\\
\hline
\textit{Static} & 256 & 4 & 3 & 2.825\\
\textit{Dynamic} & 256 & 4 & 3 & \underline{0.745}\\
\hline
\textit{Static} & 512 & 3 & 2 & 1.851\\
\textit{Dynamic} & 512 & 3 & 2 &  \underline{0.677}\\
\hline
\end{tabular}
\end{center}
\caption{Ablation study for our model using the aforementioned sequences at different resolutions. It shows how the dynamic approach performs consistently better than the static one.
\label{tab:ablation-time}}
\end{table}


\section{Conclusions}
In this paper, we presented a fully convolutional graph-based approach for time-varying point clouds upscaling using a novel and different approach with respect to most of the state-of-the-art models. 
Our proposed method is comparable with state-of-the-art solutions in terms of upsampling performance but it has a much lighter architecture that allows the implementation on edge devices with limited computational capabilities. 

As a future development this type of application could be implemented as an update for older LiDAR devices or to allow faster 3D point cloud streaming by only transmitting/sampling a subset of the original points.

While out method tackles the problem in a different way bringing some advantages it still has some limitations and drawbacks:
\begin{itemize}
    \item Training time and memory footprint. Not relying on an encoder-decoder model implies having the whole point cloud at every stage of the network in memory, this slows down training and poses some limitations in the number of input frames;
    \item  Results for the reconstruction accuracy are comparable with those reported in the state-of-the-art, though a bit lower. 
\end{itemize}

\section{Acknowledgments}
This work was supported by the European Commission under European Horizon 2020 Programme, grant number 951911—AI4Media.





\bibliographystyle{ACM-Reference-Format}
\bibliography{sample-base}

\end{document}